\documentclass[conference]{IEEEtran}
\IEEEoverridecommandlockouts

\usepackage{cite}
\usepackage{amsmath,amssymb,amsfonts,amsthm}
\usepackage{algorithmic}
\usepackage{graphicx}
\usepackage{textcomp}
\usepackage{xcolor}
\usepackage{multirow}
\usepackage{stfloats}
\usepackage{numprint}
\usepackage{subfig}
\usepackage{enumitem}

\def\BibTeX{{\rm B\kern-.05em{\sc i\kern-.025em b}\kern-.08em
    T\kern-.1667em\lower.7ex\hbox{E}\kern-.125emX}}

\usepackage{hyperref}

\newif\ifNota
\Notafalse
\newcommand{\nota}[1]{\ifNota \textcolor{red}{#1} \fi}
\newcommand{\CK}{\hspace{2mm} \checkmark \hspace{2mm} }

\newcommand{\methodsc}{\textsc{TEDD}}

\newtheoremstyle{problemstyle}
{}                
{}                
{}        
{}                
{\bfseries}       
{.}               
{\newline}               
{}                

\theoremstyle{problemstyle}\newtheorem{problem}{Problem}

\begin{document}

\title{TEDD: Robust Detection of \\ Unstable Temporal Features
\thanks{\textsuperscript{1} Work developed during an internship at Feedzai.}\thanks{\textsuperscript{2} Work developed while employed at Feedzai.}}

\author{
\IEEEauthorblockN{Ricardo Pereira\textsuperscript{1}}
\IEEEauthorblockA{\textit{DCC/FCUP University of Porto}\\
Porto, Portugal\\
rrpereira@fc.up.pt}
\and
\IEEEauthorblockN{Bruno Casal Laraña}
\IEEEauthorblockA{\textit{Feedzai} \\
Porto, Portugal\\
bruno.larana@feedzai.com}
\and
\IEEEauthorblockN{Nádia Soares\textsuperscript{2}}
\IEEEauthorblockA{\textit{Mindera}\\
Porto, Portugal \\
nadia.soares@mindera.com}
\and
\IEEEauthorblockN{Miguel Araújo}
\IEEEauthorblockA{\textit{Feedzai}\\
Porto, Portugal\\
miguel.araujo@feedzai.com}
}

\maketitle              

\begin{abstract}
When working with real-world temporal data, it is common to encounter features whose distribution is changing over time. The naive employment of Machine Learning models on this unstable data might lead to rapidly degrading performance, especially if the new distribution is much different from what was previously seen during training.
In order to cope with this problem, it is critical to automatically identify features that are changing over time. With these features detected, data scientists and other practitioners will be able to mitigate the issue (for instance, by applying data transformations), deploying more robust models that retain high performance for longer periods of time.

In this paper, we describe which temporal changes a feature should not suffer from, and propose \methodsc, a technique to a) identify when a dataset might lead to an unstable Machine Learning model and b) automatically detect which features cause such lack of robustness. In order to achieve it, we leverage a regression model to highlight which features contribute to a good prediction of an instance's timestamp.  
We compare our approach to other methods in real and synthetic data, testing their detection capability on all simple change patterns.
We show that our method: detects all types of basic changes, both for numerical and categorical features; can detect multivariate drifts; returns a comparable value measuring the amount of change of each feature; requires no parameter tuning; and is scalable both on number of features and instances of the dataset.
\end{abstract}

\section{Introduction}

The purpose of a data scientist or machine learning engineer is to build a Machine Learning model using existing data, and then apply it to new, never-before-seen instances. During this process, a common assumption underpinning most Machine Learning algorithms is that feature distributions do not change over time, a concept often called the stationary hypothesis. For instance, think about building a model to predict whether a picture contains a cat or a dog - one can somewhat safely assume that new images will share many properties with old images\footnote[3]{\ We still wouldn't want to use a ML model trained on pictures from the 1940s to evaluate 2020 images. We can see that temporal evolution is still happening, but at a much slower pace.}. However, it is quite common to encounter non-stationary environments when working in real-world applications. Consider fraud detection, churn prediction or product recommendations. These are dynamic environments where data attributes may change quickly.

We can see that whether new instances are similar to those used to train our model is a characteristic of the problem or data collection method and it is typically not under our control. This raises a problem because, if there is a change in the distribution of some features in the dataset, the function that
the model learned using the training data might be outdated when the model is deployed. When one takes into consideration typical delays between model training and model deployment (months, at some companies), then this problem is exacerbated even further.
In that case, the non-stationarity of the data will most likely result in a significant loss of the model's predictive power. This has significant real world impact, as model decay leads to higher operational costs (additional model retrainings are necessary) or, even worse, to erratic and potentially unfair behavior towards end-users.

In order to prevent this, we first need to detect that a change is taking place.
So, how can we detect features whose distribution is changing over time?
Which are the desired properties for such detection method?
What type of changes are we expected to encounter?
We present the two problems we address in this paper:

\vspace{-1mm}
\begin{problem}[Detecting drift]
\textbf{Given:} a dataset with $N$ instances, each with an associated timestamp and a set of associated attributes (features).\\ 
\textbf{Find:} a measurement of how much our data is affected by features who are not stable over time.
\end{problem}
\vspace{-2mm}
\begin{problem}[Identifying critical features]
\textbf{Given:} the same setting of Problem 1,\\
\textbf{Find:} a ranking of the features, sorted by how much they are changing over time.
\end{problem}
\vspace{-1mm}


Our contributions can be summarized as follows:
\begin{itemize}
    \item \textbf{Properties:} We identify the desired properties and key problems that a solution to this problem should address, for example the types of change that it should be able to detect for both numerical and categorical features.
    \item \textbf{Method:} We propose a novel technique called Temporal Drift Detector (\methodsc) that verifies all the above identified properties while being able to work out of the box.
    \item \textbf{Experiments:} We compare our method to other four techniques using real and synthetic data, showing that we do a better job detecting change, specially for the cases of erratic behaviour and multivariate change.
\end{itemize}

\section{Related Work}

The most common methods for detecting a change in the distribution of data are statistical tests, either parametric or non-parametric.
They can be used to compare specific statistics (like mean and variance) between two samples and test if they are significantly different.
Examples of this type of methods include the Student t-test \cite{ttest} and the Fisher f-test \cite{ftest} for the parametric group; and the Mann-Whitney U-test \cite{utest} and the Wilcoxon signed-rank test \cite{wilcoxon} for the non-parametric one.

While the tests mentioned in the previous paragraph only assert that a chosen statistic of two samples is similar, statistical distances gives us a measure of how different the distributions of those two samples really are.
Some examples of these metrics are the Kolmogorov-Smirnov test \cite{ks}, the Kullback-Leibler Divergence \cite{kl} and the Wasserstein Distance \cite{wass} (also known as the Earth Mover's Distance).

Both statistical tests and distances were meant to distinguish between distributions, so the natural way to apply them to our use case is to treat the features independently, split the dataset into several time windows and compare the distributions of the corresponding samples pairwise.
Results can then be aggregated for each pair of windows in order to measure the overall change.

Another common approach is using control charts to monitor the data as time passes. These are techniques that try to identify a point in time when the underlying distribution changed.
The Shewhart individuals control chart \cite{cchart} is a simple example, in which we use an initial sample to determine a baseline and upper and lower limits for the data, considering points outside these limits to be out of control.
The exponentially weighted moving average (EWMA) chart \cite{ewma} is a relaxed version of the Shewhart's in which we also establish a upper and lower limit, but we only deem the process out of control if the EWMA is outside these boundaries.
The cumulative sum (CUSUM) control chart \cite{cusum} is another variation of this method, in which we keep track of the sum of the deviations from the expected value and alert if it goes beyond a chosen threshold.
Lastly, the Adaptive Windowing algorithm (ADWIN) \cite{adwin} uses a sliding window approach where the size of the window is automatically updated depending on the current behaviour of the data stream, providing rigorous guarantees of its performance.

Other approaches include the generalized likelihood ratio \cite{glr}, the interception of confidence intervals \cite{ici} and methods based on novelty detection \cite{novelty}, to name a few.

Most methods mentioned above have been proposed as ways to detect covariate shift, which refers to a change in the distribution of the input features (not the label) between training set and test set when training Machine Learning models \cite{unifying}. When working with temporal data, covariate shift might be seen as a special case of the problem that we are trying to solve, as it is common to separate the available data in two separate time windows, the oldest being used for training and the most recent for testing. Therefore, a change in distribution between train and test implies a change over time.

There is a widely used ad-hoc method to detect covariate shift\footnote{We were not able to find the original source for this method, even though the method is mentioned in several blog posts and data science competitions.}, which we will call CST (covariate shift test). 
After splitting the data between train and test, we add a new label with the origin of each instance (train or test) and train a classification model to predict this new label given the value of the remaining features. The better the model performs, the greater the change is.
There are slight variations of this method: one can either train the model with one feature at a time and compare the scores to see which feature is changing the most; or train the model with all features and use a feature importance metric to discriminate them.

\section{Proposed Method}

\subsection{Important Properties}\label{sec:properties}
Which properties should a drift detection method satisfy? 

Let $\mathbf{X}$ be our $T\times N$ feature matrix, where each row corresponds to a timestamp and each column to a specific feature of our dataset. Let $\mathbf{X_i}$ represents one of its columns and $\mathbf{t}$ be the vector of timestamps associated with each instance.

We want a detection method that alerts whenever a feature has one of the following properties:

\begin{enumerate}
    \item For numerical features:
    \begin{enumerate}[label=P\arabic*.]
        \item \textit{Linear change of mean}\\
        Methods should detect simple linear changes of mean values, i.e. $E[\mathbf{X_i}] = a_i t$.
        \item \textit{Abrupt change of mean}\\
        Mean changes at a given timestamp $k$, i.e. 
        \begin{equation*}
            E[\mathbf{X_i}] = 
            \begin{cases}
                a\text{, if } t < k\\
                b\text{, if } t \geq k
            \end{cases}
        \end{equation*}
        \item \textit{Linear change of variance}\\
        Variance linearly changes, i.e. $Var[\mathbf{X_i}] = a_i t$.
        \item \textit{Abrupt change of variance}\\
        Variance changes at a given timestamp $k$, i.e. 
        \begin{equation*}
            Var[\mathbf{X_i}] = 
            \begin{cases}
                a\text{, if } t < k\\
                b\text{, if } t \geq k
            \end{cases}
        \end{equation*}
    \end{enumerate}
    \item For categorical features:
    \begin{enumerate}[label=P\arabic*.]
        \setcounter{enumii}{4}
        \item \textit{Change of relative frequency}\\
        Methods should detect when the relative frequency of values in a categorical variable changes significantly, i.e., when the probability of seeing a specific categorical value is not constant when considering time:
        \begin{equation*}
            P(\mathbf{X_i} = k|\mathbf{t}) \neq a
        \end{equation*}
        \item \textit{Change of domain}
        If a new categorical value is suddenly introduced, i.e.
        \begin{equation*}
            P(\mathbf{X_i} = a|\mathbf{t}) = \begin{cases}
                0\text{, if } t < k\\
                b\text{, if } t \geq k
            \end{cases}
        \end{equation*}
    \end{enumerate}
    \pagebreak
    \item General:
    \begin{enumerate}[label=P\arabic*.]
        \setcounter{enumii}{6}
        \item \textit{Erratic behaviour}\\
        Whenever, for some periods, feature values are not available or misreported (e.g., values are reported as missing, null or 0 for those periods).\\
        Let $L = [(a_1, b_1), ..., (a_k, b_k)]$ be a series of $k$ non-periodic non-overlapping intervals and $\mathbf{X_i^e}$ be the subset of entries of $\mathbf{X_i}$ that occurred in one of the intervals defined in $L$ and $\mathbf{X_i^{n}} = \mathbf{X_i} \backslash \mathbf{X_i^e}$ represent the remaining observations during normal behavior. Then we consider a feature to be erratic whenever
        \begin{equation*}
            \mathbf{X_i^e} = 0 \text{ and } \mathbf{X_i^n} \neq 0
        \end{equation*}
        
        \item \textit{Multivariate change}\\
        The distribution of every feature is stable over time, i.e. $P(\mathbf{X_i}|\mathbf{t}) = P(\mathbf{X_i})\ \forall\ i$, but there are multivariate dependencies on time, i.e. $P(\mathbf{X}|\mathbf{t}) \neq P(\mathbf{X})$.
    \end{enumerate}
\end{enumerate}

Additionally, a detection method should also be:
\begin{enumerate}
    \item[] \begin{enumerate}[label=P\arabic*.]
        \setcounter{enumii}{8}
        \item \textit{Linearly scalable}\\
        Linear scalabilty both in the number of instances and number of features is important given the size of common datasets.
        
        \item \textit{Parameter-free}\\
        As drift detection is ideally performed as a sanity-check at the beginning of a Data Science pipeline, one shouldn't expect users to spend significant time tuning parameters.
    \end{enumerate}
\end{enumerate}

\subsection{Problem 1 - Detecting drift}
We'll start by identifying how one can measure how much our dataset is affected by unstable features.

Let $x$ be a $p-$dimensional random variable that follows the distribution of the feature vector of the dataset. Saying that the distribution of features does not change across time is the same as to say that $\forall t_i, t_j: P(x|t_i)=P(x|t_j)$, i.e., knowing the time at which this instance was observed has no influence on our probability distribution.

If the probability distribution of $x$ is the same for every value of $t$, we can say that $P(x|t)=P(x)$.
From there, we can conclude by trivial application of Bayes' theorem
\begin{equation*}
    P(x|t) = P(x) \Leftrightarrow \frac{P(t|x)P(x)}{P(t)} = P(x)
        \Leftrightarrow P(t|x) = P(t)
\end{equation*}

Therefore, we can detect if there is a change in the distribution of our features over time by verifying if knowledge of our feature vector affects the probability distribution over the timestamps in our dataset.

Now, if we train a regression model $M$ to predict timestamp based on the feature vector, it will learn to approximate $P(t|x)$, usually by minimizing the mean squared error (MSE).
If the feature vector is not changing over time, then $M$ is trying to make a prediction given random noise. In order to minimize the MSE, $M$ will converge to predicting the mean value of the timestamp vector.
Anything better can only be achieved if M was able use the value of $x$ to make a better prediction of $t$, which would contradict our initial hypothesis.

We propose the use of the coefficient of determination of our regression model, which indicates how much better $M$ is than a baseline model always predicting the mean, as a measure for how much our dataset is affected by features which are changing over time.
\begin{equation*}
R^2 = 1-\frac{\sum_{i=1}^{N}(t_i-M(X_i))^2}{\sum_{i=1}^{N}(t_i-\overline{t})^2}
\label{eq2}
\end{equation*}

If there is no change in the distribution and $M$ is similar to the baseline model ($\forall$ instances $i$, $M(x_i)\approx \overline{t_i}$), $R^2$ will be very close to $0$.
Conversely, the closer this value is to $1$, the greater the change in distribution needs to be.

\subsection{Problem 2 - Ranking features by amount of change}

Now that we know how to measure how much influence time has over the distribution of our features, we would like to rank them according to their impact.

We propose the use of standard feature importance techniques to rank all features by their contribution to the prediction of timestamp.

Given that we measure the amount of change in the distribution of features by the reduction of the mean-squared error in relation to the baseline model and that the feature importance measured as described above serves as a way to give credit to each feature proportionally to the its contribution towards that reduction, this aligns directly with what we needed.

\subsection{Implementation}

When selecting a specific Machine Learning technique to train model $M$, we need to select an algorithm that allows us to achieve all properties in sub-section \ref{sec:properties}. For instance, one should use neither a linear model (as it won't detect multivariate changes) nor SVM (as they typically do not offer linear scalability). Other techniques are not a good choice as they require significant parameter tuning, e.g. Deep Learning architectures and k-Nearest Neighbors distance functions.

We choose Random Forest as the regressor model of \methodsc\ because it is a robust method that can handle any type of tabular data. As a decision tree based model, its standard feature importance metrics have the desired properties specified above and most implementations already have a built-in function for that purpose.
In this case, the feature importance is measured by the normalized total reduction of the cost function of the nodes that split by that feature.

Also, regarding parameter tuning, we find that using common default values for the model's parameters yields consistent results.
Namely, we used 100 estimators (trees) with a maximum of depth of 32 and the square root of the number of features to search for the best split.

\section{Results}

We conduct several experiments on synthetic and real data to answer the following questions:

\begin{enumerate}[label=\textbf{Q\arabic*.}]
\item Can \methodsc\ alert whenever a feature triggers one of the desired properties?
\item Is \methodsc\ linearly scalable?
\end{enumerate}

\begin{figure}[t]
\centering
\subfloat{\includegraphics[width=.45\textwidth]{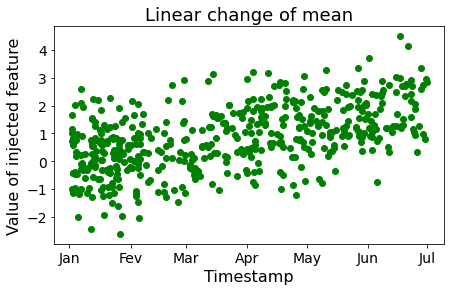}}
\hfil
\subfloat{\includegraphics[width=.45\textwidth,height=4.5cm]{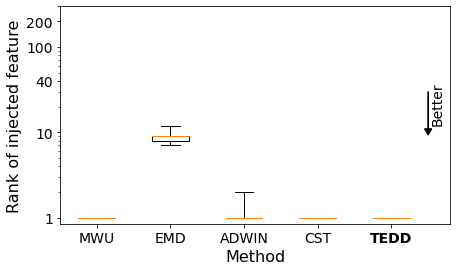}}
\caption{Numerical feature with a linear change of mean. We show a scatter plot of a sample of it (top) and a box plot of its ranking by each method over 20 runs (bottom).}
\label{fig_gradual_mean}
\end{figure}

In the following, we start by introducing the experimental setup and datasets used, before detailing all experiments conducted. We compare \methodsc\ with different types of state of the art methods and show that we are able to pick up every relevant type of change (both for numerical and categorical variables) and do it in a way that is scalable (both on the number of features and instances).

\begin{figure}[t]
\centering
\subfloat{\includegraphics[width=.45\textwidth]{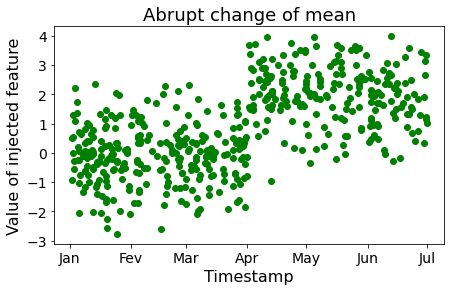}}
\hfil
\subfloat{}{\includegraphics[width=.45\textwidth]{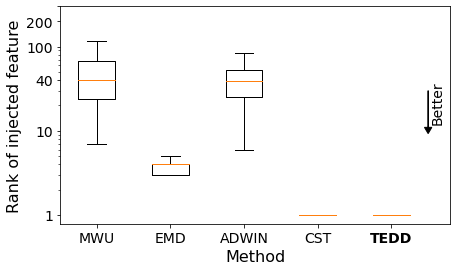}}
\caption{Numerical feature with an abrupt change of mean. We show a scatter plot of a sample of it (top) and a box plot of its ranking by each method over 20 runs (bottom).}
\label{fig_abrupt_mean}
\end{figure}

\subsection{Experimental setup}

We selected one method from each type mentioned in the Related Work section: one statistical test, one statistical distance, one control chart technique and the standard Covariate Shift Test (CST).

We chose the Mann-Whitney U-test (MWU) as the representative of the statistical tests group since it is non-parametric and used for independent samples.
From the statistical distance group we chose the Earth Mover's Distance (EMD) because its definition is aligned with the problem that we have at hand.
From the control chart group we chose the Adaptive Windowing algorithm (ADWIN) because it is state of the art control chart and has an available implementation.
Lastly, we decided to use a random forest classification model for the CST, training a single model with all features and ranking them according to feature importance, since it was the most similar option to our method.

We tested the ability of each of aforementioned change patterns separately by injecting a controlled feature with that type of change in the dataset.
We then run all the methods we are comparing and check in which position of the ranking the injected feature appears.
As we need to sample the original dataset (some methods don't scale) and some models are not deterministic, each experiment was run \numprint{20} times with different seeds.
For every experiment, we present a visualization of the injected feature and a box plot of the ranking of said feature for each method.
We also mention what was the change in the $R^2$ of the regression model after injecting each feature.

\begin{figure}[htbp]
\centering
\subfloat{\includegraphics[width=.45\textwidth]{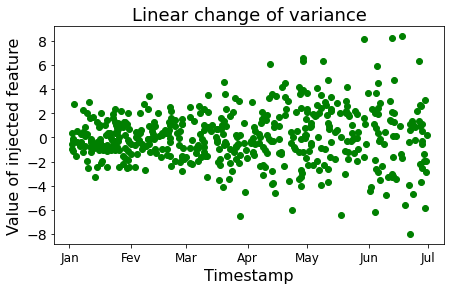}}
\hfil
\subfloat{\includegraphics[width=.45\textwidth]{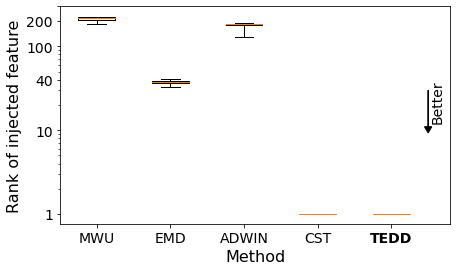}}
\caption{Numerical feature with a linear change of variance. We show a scatter plot of a sample of it (top) and a box plot of its ranking by each method over 20 runs (bottom).}
\label{fig_gradual_variance}
\end{figure}

\subsection{Datasets}

We used the IEEE-CIS Fraud Detection dataset \cite{ieee} for our injection experiments.
This dataset includes information of \numprint{590540} card transactions, each with \numprint{434} attributes.
It includes features such as the amount being paid, the moment it took place, billing address and credit card information, to name a few.
The goal is to predict the binary label, \texttt{isFraud}, in which the positive class corresponds to fraudulent transactions and the negative to legitimate ones. 
As some features were empty or constant, a simple data cleaning step was performed. After imputing missing values and dropping features that were at least \numprint{95}\% constant, we end up with a clean dataset with \numprint{590540} instances and \numprint{222} features.

Regarding time information, the \texttt{TransactionDT} feature ``is a timedelta from a given reference datetime (not an actual timestamp)".
Given the range of values, it seems to be in seconds, spanning for near six months.
For plotting purposes, we will assume it started on 2019/01/01.
The dataset includes a feature named \texttt{TransactionID} which is an indexing numerical feature used to join tables.
However, as it was assigned in chronological order, it is highly correlated with timestamp.
These are the only two features with high dependency with time and every method consistently ranked them at the top.
Since their ranking was obvious, we decided to discard these two features in order to simply the results.
This way, the optimal result for all experiments would be ranking the injected feature as changing the most (rank \numprint{1}).

After removing these two features, the regression model of \methodsc\ got an $R^2$ score of \numprint{0.44}.
For every experiment we will report how much this value changed after the addiction of the injected feature.

\begin{figure}[htbp]
\centering
\subfloat{\includegraphics[width=.45\textwidth]{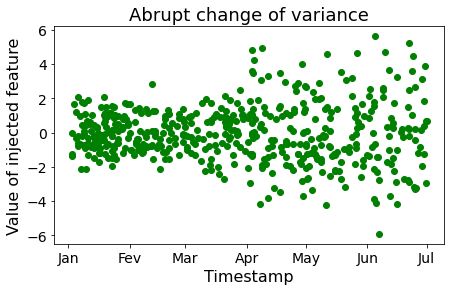}}
\hfil
\subfloat{\includegraphics[width=.45\textwidth]{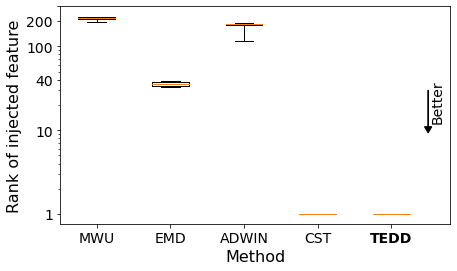}}
\caption{Numerical feature with an abrupt change of variance. We show a scatter plot of a sample of it (top) and a box plot of its ranking by each method over 20 runs (bottom).}
\label{fig_abrupt_variance}
\end{figure}

To test the scalability in the number of instances, we used the WSDM - KKBox's Churn Prediction dataset \cite{churn}, which is a collection of \numprint{20148758} entries with \numprint{22} features, with user information like membership status, statistics of service usage and payment data.
The goal is to predict the binary label \texttt{isChurn}, defined as ``whether the user did not continue the subscription within 30 days of expiration".

\subsection*{P1. Linear change of mean}

The injected numerical feature with a gradual change of mean follows a normal $\mathcal{N}(\mu,\sigma^2)$ distribution whose mean grows linearly from \numprint{0} at the beginning of the dataset to \numprint{2} at the end, while keeping the standard deviation constant at \numprint{1} (see Figure~\ref{fig_gradual_mean}).
This feature resulted in an increase of the $R^2$ score of the regression model by \numprint{0.07}.
As we can see, most methods did a nearly perfect job ranking this feature, the only exception being EMD which still ranked it in the top 10 \numprint{95}\% of the time.

\begin{figure}[htbp]
\centering
\subfloat{\includegraphics[width=.45\textwidth]{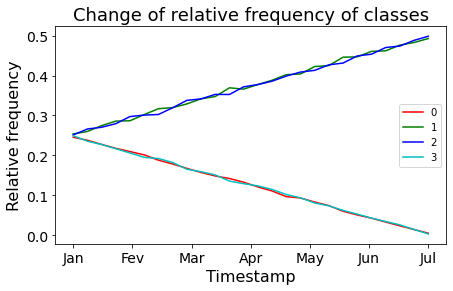}}
\hfil
\subfloat{\includegraphics[width=.45\textwidth,height=4.4cm]{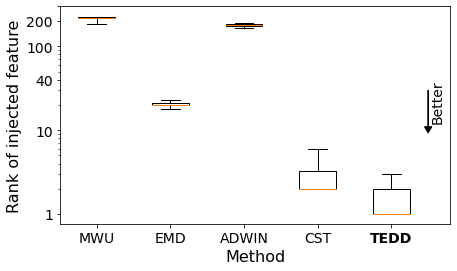}}
\caption{Categorical feature with a change in the relative frequency of classes. We show a line plot of their relative frequency each week (top) and a box plot of its ranking by each method over 20 runs (bottom).}
\label{fig_rel_freq}
\end{figure}

\vspace{-1mm}
\subsection*{P2. Abrupt change of mean}

The injected numerical feature with an abrupt change of mean follows a normal distribution $\mathcal{N}(0,1)$ during the first three months of the data, changing to $\mathcal{N}(2,1)$ afterwards (see Figure~\ref{fig_abrupt_mean}).
This feature resulted in an increase of the $R^2$ score of the regression model by \numprint{0.09}.
As we can see, both \methodsc\ and CST were able to rank it perfectly all \numprint{20} runs and EMD kept it consistently on top 10, while the other two methods were very inconsistent and ranked it much lower.

\vspace{-1mm}
\subsection*{P3. Linear change of variance}

The injected numerical feature with a gradual change of variance (a case of heteroscedasticity) follows a normal $\mathcal{N}(\mu,\sigma^2)$ distribution whose standard deviation grows linearly from \numprint{1} at the beginning of the dataset to \numprint{3} at the end, while keeping the mean constant at \numprint{0} (see Figure~\ref{fig_gradual_variance}).
This feature resulted in an increase of the $R^2$ score of the regression model by \numprint{0.03}.
As we can see, \methodsc\ and CST were able to keep a perfect score while the rest of the methods ranked it consistently low.

\begin{figure}[htbp]
\centering
\subfloat{\includegraphics[width=.45\textwidth]{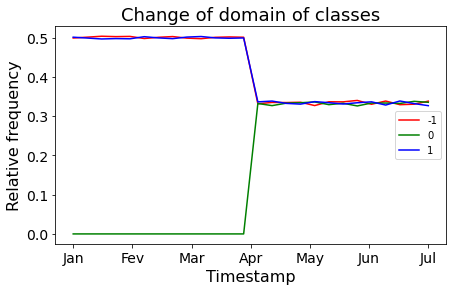}}
\hfil
\subfloat{\includegraphics[width=.45\textwidth,height=4.4cm]{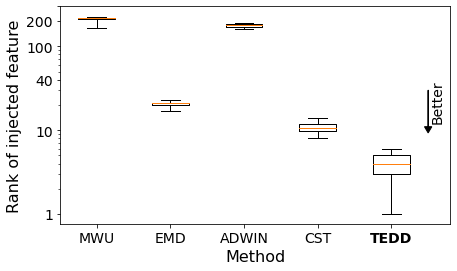}}
\caption{Categorical feature with a change in the domain. We show a line plot of the relative frequency of its classes each week (top) and a box plot of its ranking by each method over 20 runs (bottom).}
\label{fig_domain}
\end{figure}

\subsection*{P4. Abrupt change of variance}

The injected numerical feature with an abrupt change of variance (another case of heteroscedasticity) follows a normal distribution $\mathcal{N}(0,1)$ during the first three months of the data, changing to $\mathcal{N}(0,2)$ after that (see Figure~\ref{fig_abrupt_variance}).
This feature resulted in an increase of the $R^2$ score of the regression model by \numprint{0.03}.
As we can see, the results were very similar to the linear change of variance, with \methodsc\ and CST having a perfect score while the rest of the methods performing poorly.

\subsection*{P5. Change of relative frequency}

The injected categorical feature whose classes change their relative frequency has \numprint{4} unique values.
They are uniformly distributed at the beginning of the dataset and gradually change so that, by the end of the dataset, two of them have a \numprint{50}\% relative frequency and the other two have \numprint{0}\% (see Figure~\ref{fig_rel_freq}).
Since some of the methods need categorical features to be encoded, we will use the values of \numprint{1} and \numprint{2} for the first pair and \numprint{0} and \numprint{3} for the second.
This way, the numerical properties (mean and variance) of the feature don't change drastically throughout the dataset, which means that if a change is detected it will not be because of already tested properties from the method.
This feature resulted in an increase of the $R^2$ score of the regression model by \numprint{0.02}.
As we can see, \methodsc\ ranked this feature very close to the top, followed by CST.
EMD detected some drift but didn't rank it very high and the other two didn't seem to notice any drift at all.

\begin{figure}[htbp]
\centering
\subfloat{\includegraphics[width=.45\textwidth]{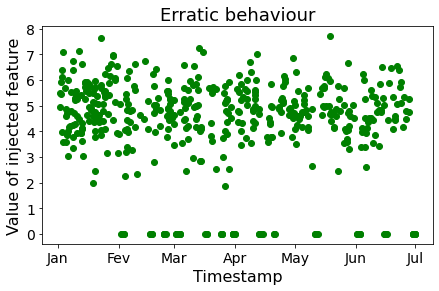}}
\hfil
\subfloat{\includegraphics[width=.45\textwidth,height=5cm]{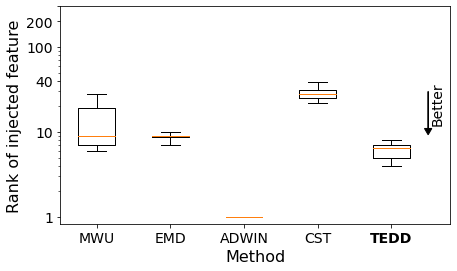}}
\caption{Numerical feature with erratic behaviour. We show a scatter plot of a sample of it (top) and a box plot of its ranking by each method over 20 runs (bottom).}
\label{fig_erratic}
\end{figure}

\subsection*{P6. Change of domain}

The injected categorical feature whose domain changes in the dataset has two unique values during the first three months of the dataset and three unique values after that, always being uniformly distributed according to the current domain.
To keep the numerical properties under control as we mentioned above, the domain of the first half will be $\{-1,1\}$ and $\{-1,0,1\}$ for the second half (see Figure~\ref{fig_domain}).
This feature resulted in an increase of the $R^2$ score of the regression model by \numprint{0.02}.
As we can see, \methodsc\ could occasionally rank it near top 1 and consistently below top 10, CST and EMD gave acceptable results between rank 10 and 20 and MWU and ADWIN not being able to detect any change.

\begin{figure}[htbp]
\centering
\subfloat{\includegraphics[width=.45\textwidth]{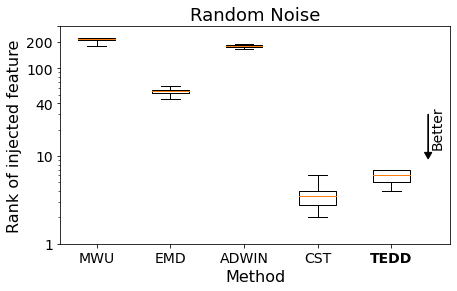}}
\hfil
\subfloat{\includegraphics[width=.45\textwidth]{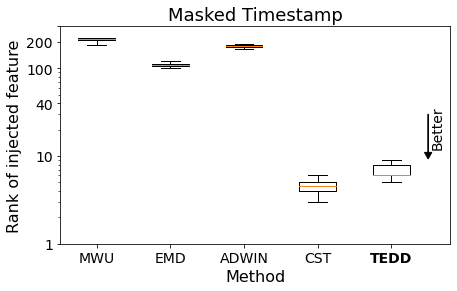}}
\caption{Rankings of injected features for multivariate change detection experiment.}
\label{fig_multivariate}
\end{figure}

\subsection*{P7. Erratic behaviour}

To simulate a type of erratic behaviour, we started with a normal distributed feature following $\mathcal{N}(5,1)$ and each weekend it had a \numprint{50}\% chance of being "missing" and imputed with value \numprint{0} (see Figure~\ref{fig_erratic}).
This feature resulted in an increase of the $R^2$ score of the regression model by \numprint{0.01}.
As we can see, only ADWIN was able to perfectly rank it all \numprint{20} runs, \methodsc\ and EMD performed decently and the other two didn't rank it very high.

\begin{table*}[b]
\center
\caption{Comparison of state of the art methods}
\begin{tabular}{|l|c|c|c|c|c|c|c|c|c|c|}
\hline
                            & Abrupt & Gradual & Mean & Variance & RelFreq & Domain & Erratic & Multi & Param-free & Scal \\ \hline
Statistical Tests (MWU)     &        & ok      & ok   &          &         &        &         &       &      & ok   \\ \hline
Statistical Distances (EMD) & ok     &         & \CK  &          &         &        & ok      &       &      & ok   \\ \hline
Control Charts (ADWIN)      &        & ok      & ok   &          &         &        & \CK     &       & \CK  &      \\ \hline
CST                         & \CK    & \CK     & \CK  & \CK      & \CK     & ok     &         & \CK   &      & \CK  \\ \hline
\methodsc                        & \CK    & \CK     & \CK  & \CK      & \CK     & \CK    & \CK     & \CK   & \CK  & \CK  \\ \hline
\end{tabular}
\label{tab_methods}
\end{table*}

\subsection*{P8. Multivariate changes}

To test the ability to detect multivariate changes (in which two or more variables are changing together), we need two variables to have a dependency with timestamp only when viewed together.
For this experiment, we injected two features: the first being random noise, a numerical feature uniformly distributed between $10^2$ and $10^4$; and the second being the product of the first feature by the timestamp at each instance.
Since timestamp only increases by \numprint{1}\% between its minimum and maximum values and the random noise feature spans over different orders of magnitude, there will be little correlation between the second feature and the original timestamp.
These features resulted in no significant change of the $R^2$ score of the regression model.
As we can see in Figure~\ref{fig_multivariate}, as expected, only \methodsc\ and CST were able to catch this multivariate change, consistently ranking both injected features on top 10 while the other methods considered them just random noise.

\begin{figure}[htbp]
\centering
\subfloat{\includegraphics[width=.45\textwidth]{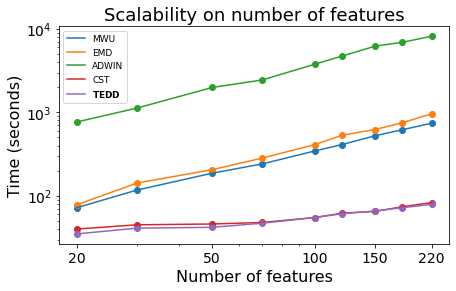}}
\hfil
\subfloat{\includegraphics[width=.45\textwidth]{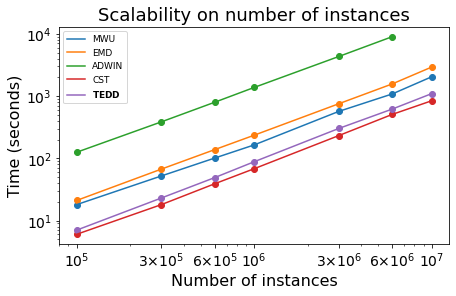}}
\caption{Scalability test on number of features (top) and instances (bottom) for each method. Notice the logarithmic scale on both axis.}
\label{fig_scale}
\end{figure}

\vspace{1mm}
\subsection*{P9. Scalability}
\vspace{-5mm}
For the scalability on the number of features, we used the same IEEE Fraud dataset as before, without any imputed features, while varying the number of features in the dataset by random sampling.

For the scalability on the number of instances, we used the KKBox Churn dataset as mentioned above, varying the number of entries in the dataset by random sampling.

Figure~\ref{fig_scale} shows the average running time after 3 runs. Notice that the plot is in log-log scale.
As we can see, even though all methods seem to grow linearly, both \methodsc\ and CST are consistently faster, followed by MWU and EMD with ADWIN significantly slower.
On the first plot we can even see an order of magnitude of difference between those three groups, with the model based methods being \numprint{10} times faster than the statistical methods and \numprint{100} times faster than ADWIN.

\subsection{Final remarks}

We summarize the results from every experiment on Table~\ref{tab_methods}.
We use three levels of performance: consistently ranking injected features near the top has a check mark, ranking them near top 10 is considered ok and below that is left blank.
However, we should note that since all experiments were performed independently, it is reasonable to believe that some injected features were easier to detect than others.

In most experiments, \methodsc\ and CST performed much better than the other three methods. The main differences
between these two techniques was on the domain change and erratic behaviour experiment, in which \methodsc\ performed better. 
The remaining methods struggle to detect changes of the variance on numerical features, changes on categorical features and, of course, multivariate change.
A notable exception was the erratic behaviour experiment where ADWIN was able to detect it perfectly while the CST performed particularly worse than on the rest of the experiments.

Regarding the $R^2$ metric, we can see that it clearly aligns with the ability of \methodsc\ to detect change: for higher values of $R^2$, it was certain that the injected feature was the one changing the most; for smaller values of $R^2$, it was more difficult to distinguish it from the rest of the features.


\section{Conclusion}

In this paper, we address the problem of detecting if features of a temporal dataset change their distribution over time.
Our contributions are:

\begin{itemize}
    \item \textbf{Properties:} We identify the desired properties that a solution to this problem should have, namely: detecting all types of basic change, both for numerical and categorical features; detecting erratic behaviours and multivariate change; and be linearly scalable and parameter-free.
    \item \textbf{Method:} We propose a new method, \methodsc\ , that verifies all the desired properties using a random forest regression model trained to predict the timestamp of instances of the dataset to rank the features that are changing the most according to their feature importance.
    \item \textbf{Experiments:} We compare our method to other  four techniques using real and synthetic data, showing that we do a  better job detecting change, specially for the cases of erratic behaviour and multivariate change.
\end{itemize}


\begin{thebibliography}{00}
\bibitem{ttest} R. L. Iman and W. J. Conover, A Modern Approach to Statistics, New York: Wiley, 1983
\bibitem{ftest} C.M. Douglas, Introduction to Statistical Quality Control, 5th ed., John Wiley \& Sons, USA, 2007
\bibitem{utest} \href{https://www.jstor.org/stable/2236101}{H. B. Mann and D. R. Whitney, ``On a Test of Whether one of Two Random Variables is Stochastically Larger than the Other", The Annals of Mathematical Statistics, vol. 18, no. 1, 1947, pp. 50--60.}
\bibitem{wilcoxon} \href{https://www.jstor.org/stable/3001968}{F. Wilcoxon, ``Individual Comparisons by Ranking Methods”, Biometrics Bulletin, vol. 1, no. 6, 1945, pp. 80--83.}
\bibitem{ks} G. W. Snedecor and W. G. Cochran, Statistical Methods, 8th ed., Iowa State
University Press, USA, 1989
\bibitem{kl} \href{https://www.jstor.org/stable/2236703}{S. Kullback and R. A. Leibler. “On Information and Sufficiency”, The Annals of Mathematical Statistics, vol. 22, no. 1, 1951, pp. 79--86.}
\bibitem{wass} C. Villani,  Topics in optimal transportation, Graduate Studies in Mathematics, vol. 58, American Mathematical Society, 2003.
\bibitem{cchart} \href{https://doi.org/10.1007/978-3-662-11789-7_9}{M. K. Hart and R. F. Hart, ``Shewhart Control Charts for Individuals with Time-Ordered Data", Frontiers in Statistical Quality Control, vol. 4, Heidelberg, 1992.}
\bibitem{ewma} \href{https://doi.org/10.1016/j.patcog.2014.07.028}{H. Raza, G. Prasad and Y. Li, ``EWMA model based shift-detection methods for detecting covariate shifts in non-stationary environments", Pattern Recognition, vol. 48, 2015, pp. 659--669.}
\bibitem{cusum} \href{https://www.jstor.org/stable/2333009}{E. S. Page, ``Continuous inspection schemes”, Biometrika, vol. 41, 1954, pp. 100--115.}
\bibitem{adwin} \href{https://www.cs.upc.edu/~gavalda/papers/adwin06.pdf}{A. Bifet and R. Gavaldà, ``Learning from Time-Changing Data with Adaptive Windowing", Society for Industrial and Applied Mathematics, 2007.}
\bibitem{glr} \href{https://ieeexplore.ieee.org/abstract/document/1101146}{A. Willsky and H. Jones, ``A generalized likelihood ratio approach to the detection and estimation of jumps in linear systems", IEEE Transactions on Automatic control 21.1, pp. 108--112, 1976.}
\bibitem{ici} \href{https://doi.org/10.1016/j.neunet.2011.05.012}{C. Alippi, G. Boracchi and M. Roveri, ``A just-in-time adaptive classification system based on the intersection of confidence intervals rule", Neural Networks 24.8, pp. 791--800, 2011.}
\bibitem{novelty} \href{https://doi.org/10.1016/j.sigpro.2003.07.018}{M. Markou and S. Singh, ``Novelty detection: a review – Part 1: statistical approaches", Signal Processing, 2003.}
\bibitem{unifying} \href{https://doi.org/10.1016/j.patcog.2011.06.019}{Moreno-Torres, Raeder, Alaiz-Rodriguez, Chawla and Herrera, ``A unifying view on dataset shift in classification", Pattern recognition 45.1, 2012, pp. 521-530.}
\bibitem{ieee} \href{https://www.kaggle.com/c/ieee-fraud-detection/}{IEEE-CIS and Vesta. (2019 July). IEEE-CIS Fraud Detection. Retrieved on December 2019 from https://www.kaggle.com/c/ieee-fraud-detection/}
\bibitem{churn} \href{https://www.kaggle.com/c/kkbox-churn-prediction-challenge/}{ACM WSDM and KKBox. (2017 September). WSDM - KKBox's Churn Prediction Challenge. Retrieved on December 2019 from https://www.kaggle.com/c/kkbox-churn-prediction-challenge/}
\end{thebibliography}
\end{document}